# ALOHA: Empowering Multilingual Agent for University Orientation with Hierarchical Retrieval

**Mingxu Tao**[1,3]*, **Bowen Tang**[1,2]*, **Mingxuan Ma**[1,2], **Yining Zhang**[2],
**Hourun Li**[1,2], **Feifan Wen**[1,4], **Hao Ma**[1], **Jia Yang**[1] ✉

[1]Computer Center, Peking University
[2]School of Computer Science, Peking University
[3]Center for Data Science, Peking University
[4]School of Electronics Engineering and Computer Science, Peking University
{thomastao,tbwtbw,yangj}@pku.edu.cn

## Abstract

The rise of Large Language Models (LLMs) revolutionizes information retrieval, allowing users to obtain required answers through complex instructions within conversations. However, publicly available services remain inadequate in addressing the needs of faculty and students to search campus-specific information. It is primarily due to the LLM's lack of domain-specific knowledge and the limitation of search engines in supporting multilingual and timely scenarios. To tackle these challenges, we introduce **ALOHA**, a multilingual agent enhanced by hierarchical retrieval for university orientation. We also integrate external APIs into the front-end interface to provide interactive service. The human evaluation and case study show our proposed system has strong capabilities to yield **correct**, **timely**, and **user-friendly** responses to the queries **in multiple languages**, surpassing commercial chatbots and search engines. The system has been deployed and has provided service for more than 12,000 people.

## 1 Introduction

The advancement of Large Language Models (LLMs) greatly changes the methods of information retrieval. By leveraging the powerful reasoning and generative capabilities of LLMs (Llama Team, 2024; OpenAI, 2024), users can input complex and descriptive instructions to obtain required information from conversations, which results in LLMs replacing traditional search engines in consultation scenarios.

Despite the impressive performance of LLMs, they can still make mistakes due to their lack of domain-specific knowledge. Previous works adopt retrieval-augmented generation (RAG) to refrain from hallucinated responses in the domain of medicine (Bao et al., 2023) and law (Hu et al., 2024). Commercial service such as Gemini (Gemini Team, 2024) also enhances LLMs with web search to provide reliable outputs in general domain. However, these solutions still cannot address the challenges faced in developing consulting agent for university campus.

As a large community engaged in education, research, and daily life, the university can encompass a wide variety of information sources. Faculty and students, especially the freshmen, may fail to determine where to search for information, highlighting the urgent need of consulting agent for campus. The agent should be capable to select evidence from the documents with diverse formats, which is different from retrieving medical guidelines or law articles. Furthermore, the information on campus can be frequently updated, and users have high demands of timeliness. However, search engines often fail to provide the latest web pages. Additionally, we hope the agent can be especially beneficial for the orientation of new students. The agent needs the capabilities to serve people who speak distinct languages, while search engines also demonstrate poor performance to provide cross-lingual results. As an application used daily by tens of thousands of people, we also need to ensure the agent supports human-centric interaction.

To address the issues mentioned above, we propose **ALOHA**, an <u>A</u>utomatic Multi-<u>L</u>ingual Agent for University <u>O</u>rientation with <u>H</u>ierarchical Retriev<u>a</u>l.

We collect a vast number of campus-specific documents and divide them into distinct subsets according to their structure and granularity. (1) To enhance the **correctness** of agent response, we employ hierarchical retrieval (Wu et al., 2023; Chen et al., 2024b) to find precise evidence. (2) We also prompt the LLM to be aware of the timestamp of each evidence for **timeliness**. (3) We introduce language identification (Joulin et al., 2016b) and machine translation (NLLB Team et al., 2022; Raffel et al., 2020) to support **multilingual** service. (4) We integrate external APIs through tool planning (Patil et al., 2023; Tang et al., 2023). The hyperlinks of external service are illustrated on a **user-friendly** interface.

---

*Equal contribution.

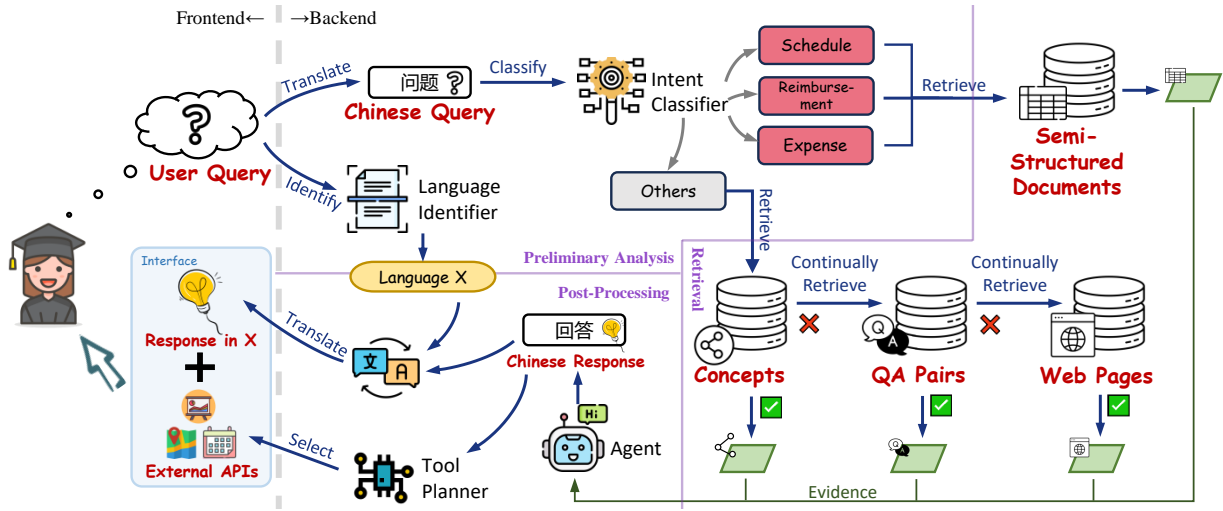

Figure 1: Overview of ALOHA, consisting of three stages: Preliminary Analysis, Retrieval, and Post-Processing.

## 2 Overview of ALOHA

Our system, ALOHA, provides interactive information consultation services for users across multiple languages. We divide the pipeline of ALOHA into three stages: **preliminary analysis** of user queries, **hierarchical retrieval**, and **post-processing** of responses. We illustrate the workflow of our system in Figure 1. More demonstrations of the interface are shown in Appendix A.

**Preliminary Analysis of Query** ALOHA can provide service to students and staff who communicate in distinct languages. Therefore, we first employ FastText (Joulin et al., 2016a) to identify the language of the query. We then translate the query into Chinese if it is in other languages, which can enhance the recall of retrieval, since almost all our documents are in Chinese.

ALOHA also conducts a pre-analysis of user intent. We find that the queries regarding schedules and financial standards typically require retrieval of semi-structured documents (i.e., containing tabular data) for accurate information. In contrast, other queries may need merely plain-text evidence. Thus, we finetune Qwen2-1.5B-Instruct (Yang et al., 2024) for user intent classification.

**Hierarchical Retrieval** To mitigate the lack of campus-specific knowledge in LLMs, we construct a collection of semi-structured and unstructured documents for retrieval-augmented generation. ALOHA can select evidence from specific subsets of documents based on the results of intent classification. We further divide the unstructured documents into three categories: concepts, question-answer pairs, and filtered web pages. Our system conducts retrieval sequentially across these categories. Since inaccurate documents may lead to hallucinations, if the system acquires evidence from a certain category, it will skip the subsequent pipelines of retrieval.

**Post-Processing of Response** We use gpt-4o to generate a response grounded in the retrieved evidence. We then translate the response into the language of original query, ensuring the language alignment with the user's input.

We also equip ALOHA with the capabilities to invoke external applications. For example, when a user requests information about a building, ALOHA can provide links that redirect to online map applications, enabling seamless access to the navigation service. Following previous work (Hsieh et al., 2023), we list the description of each tool, and gpt-4o can determine whether to use external tools and which tool should be invoked. See more details about tool usage in Appendix A.

## 3 Implementation of ALOHA

ALOHA leverages campus-specific knowledge to augment the generation of LLMs, enhancing the **correctness** and **timeliness** of the responses. However, it is challenging to retrieve accurate evidence from a collection of documents which vary in structure, granularity, and quality.

Preliminarily, we find the queries with a specific intent usually require the documents which share similar characteristics. For example, the documents related to reimbursement are almost all semi-structured. We believe that adopting empirical rules to constrain the retrieval process to a subset of documents can enhance the accuracy of retrieval results. Thus, we implement **Intent Classification** and **Hierarchical Retrieval** as the core components of ALOHA.

## 3.1 Intent Classification

We aim to determine whether LLM should leverage tabular evidence to response to a query, since the reasoning over semi-structured documents can be more complex than over plain-texts. The tabular documents primarily belong to specific areas, including reimbursement, standard of expense, and working schedule. We empirically categorize the intents within these areas into 11 classes. See more details in Appendix B.

**Data Collection** We collect 717 seed questions and use `gpt-4o` to paraphrase them, yielding a dataset of 3,573 query-intent pairs. We hold out 20% of the instances as test set. For each query and its intent, we employ `ERNIE-4.0` to generate explanations for instruction tuning.

**Heuristic Intent Classification (HIC)** To enhance the accuracy of classification, we propose a heuristic method to constrain the range of candidate classes through semantic similarity matching between queries.

ALOHA calculates the similarity of input query and each question in the training set. We select top-$k$ instances with the highest similarity scores, thereby constraining the model to make predictions within the intent classes of these $k$ instances. Since the similar queries usually share the same intent, we believe our proposed HIC can effectively filter out the incorrect classes.

We employ `BGE-m3` (Chen et al., 2024a) to extract the embeddings of both the input and the queries in training set. We measure the similarity of two queries by the cosine distance of their representation vectors. In this work, we set $k$ to 50, and we find Recall@50 of the filtered classes can achieve 1.00.

| |
|---|
| **Concept:**<br>[*School of International Studies*]: The School of International Studies at Peking University dedicates to cultivating experts in international affairs, high-level professionals for diplomatic work. ... |
| **Question-Answer Pair:**<br>*Question:* After applying the internet account for a server, is it necessary to manually configure the IP address?<br>*Answer:* No, your server can automatically obtain the IP address via DHCP. Once the application is approved, simply restart the device. |
| **Web Page:**<br>Seminar: Biochemical and Digital Networks of Metabolites<br>Time: May 30th (Thursday) 10:00-11:00<br>Location: Room B101, Lui Che Woo Building<br>Abstract: Life activities are inseparable from the metabolic network composed of small molecular metabolites... |

Table 1: Examples of different types of documents: concept, question-answer pair, and web page.

## 3.2 Hierarchical Retrieval

Retrieving accurate campus-specific knowledge plays a vital role in generating reliable and timely responses. Hence, we propose a hierarchical pipeline to prioritize the retrieval from high-quality subsets of documents.

### 3.2.1 Construction of Document Store

We collect a variety of publicly available texts related to Peking University, including student handbooks, announcements, news articles, and other web pages. We further extract query-answer pairs from these documents, for example, the Q&A sections of student handbooks.

We convert all tabular data into Markdown format, and categorize these semi-structured documents based on their corresponding intent classes. If necessary, ALOHA selects tabular evidence only from a specific subset.

However, for queries with intents not listed in Appendix B.1, ALOHA should retrieve within the collection of unstructured documents. We regard the plain-text documents as three types: concept, question-answer pair, and web page. Table 1 shows the example of each type.

### 3.2.2 Retrieval with Parsing and Reranking

The inquiry of detailed information of a person or entity is a fundamental requirement. However, when using an AI agent, users may not input merely the person or entity. They tend to give a command, such as *Introduce the School of International Studies*. Thus, we employ LTP-4 (Che et al., 2021) to parse the query for fuzzy matching between query and concepts.

We identify the subject (S), predicate (V), and object (O) through dependency parsing. If a complex query contains other syntactic constituents or a dependent clause, we regard it as mismatching to any concept. We then filter out all words except for nouns and verbs. If the filtered query remains in the form of SV or VO, or consists of merely nouns, we regard it as a simple command. We utilize the nominal subject or object of a simple command for exact matching with the collection of concepts.

For the queries that cannot be matched to concepts, we employ dense retrieval and reranking to select evidence sequentially from question-answer pairs and web pages. We obtain the embeddings of the query and documents by BGE-m3. Then, we use ElasticSearch to retrieve the top 10 documents and rerank the documents with bge-reranker-v2-m3. To mitigate the hallucinations brought by irrelevant evidence, we exclude the documents with cosine similarity scores below 0.1.

### 3.3 Scalability

We aim to provide timely and user-friendly service for inquiries of campus-specific information. Erroneous or outdated information in the responses of the LLM may mislead users. Thus, we periodically crawl the university's websites, to ensure the document store remains up-to-date.

Due to the inconsistency in the format of web pages over different periods, we cannot automatically replace specific outdated pages. We incorporate the current time and the timestamp of each evidence documents into the input of LLM. We prompt the LLM to consider the timeliness of each evidence and avoid using outdated information.

Our system can also invoke various external tools, providing interactive service to users, such as online map navigation, classroom availability, etc. We instruct gpt-4o to determine whether to use a tool based on the generated responses. The input consists of the function, primary application, and invocation method of each tool. Thus, ALOHA can support seamless integration of new plug-in tools.

## 4 Evaluation

In this section, we conduct an automatic evaluation to examine the effectiveness of our Heuristic Intent Classification. We also use human evaluation to examine the response quality of ALOHA.

| Model | Acc. | $\Delta$Acc. |
|---|---|---|
| Qwen2-1.5B | 0.340 | – |
| Qwen2-1.5B w/ HIC | 0.419 | +0.079 |
| Qwen2-7B | 0.781 | – |
| Qwen2-7B w/ HIC | 0.894 | +0.110 |
| Qwen2-1.5B-finetuned | 0.941 | – |
| Qwen2-1.5B-finetuned w/ HIC | **0.963** | +0.022 |
| ERNIE-4.0 | 0.923 | |

Table 2: Performance of Qwen2 models and ERNIE.

### 4.1 Accuracy of Intent Classification

As we mentioned in Section 3.1, we use 80% of the collected instances to finetune Qwen2-1.5B instruct model for intent classification, and the hold-out dataset for testing. We also evaluate the performance of ERNIE-4.0 and Qwen2 instruct models without finetuning. For these models, we employ 3-shot in-context learning. Experimental results are shown in Table 2.

We can find the finetuned Qwen2-1.5B model achieves an accuracy of 94.1%, even surpassing ERNIE-4.0 by 1.8%. It shows that intent classification is not a complex task. We believe that for a single simple task, finetuning a small language model presents a more efficient and effective solution, compared to using LLMs for reasoning via in-context learning.

Compared to naïve classification, our proposed HIC can enhance both vanilla Qwen2 models and the finetuned Qwen2-1.5B, providing improvements of +2.2%~+11.0% on accuracy. We also find when augmented by HIC, the performance of open-source LLM Qwen2-7B can be comparable to ERNIE-4.0. It shows the semantic similarity matching can effectively improve the accuracy of classification by filtering out irrelevant intent classes.

### 4.2 Human Evaluation

We further examine the quality of responses yielded by ALOHA through human evaluation. We collect 100 queries of real-world users, covering various areas of information consultation, including academic development, career planning, campus life, and immediate needs. To examine the system's performance in multilingual scenarios, we translate all Chinese queries into English and French separately.

We conduct a comparative study of ALOHA against following information retrieval systems, which are publicly available on the Internet:

- **ChatGPT:** A chatbot based on gpt-4o, with-

out the ability to access external information.

- **ERNIE Search:** A chatbot based on `ERNIE-4.0`, which can dynamically invoke Baidu Search for RAG.
- **Google Search:** A search engine[1] that returns multiple web pages but does not necessarily provide textual responses.

We ask senior students to examine the correctness and timeliness of the responses generated by each system. If the output contains fabricated or outdated information, we regard it as inaccurate response. Specially, for Google Search, we evaluate whether the top 10 returned web pages can address the user's query. Table 3 shows the results.

Without access to the external knowledge, ChatGPT can provide accurate responses for merely 29% of Chinese queries, while 21% of English and French queries. We find these queries are general questions which are not restricted to a particular university, such as "*How to find an internship?*"

Comparing the results of ChatGPT and ERNIE Search, we find leveraging external documents can bring improvement of +38% accuracy on the Chinese test set. We argue that **retrieved evidence can efficiently mitigate LLM's lack of campus-specific knowledge.** However, the search engine brings less improvement (+22% accuracy) on the English set. ERNIE Search tends to use only English web pages when responding to queries in English. The quantity of English pages related to Peking University is much smaller than that of Chinese pages. Similarly, ERNIE Search refuses to provide service to all French queries. We guess **the lack of cross-lingual retrieval capabilities may result in the suboptimal performance of ERNIE Search** on non-Chinese queries.

Another limitation of search engines is their inefficiency in handling queries in the format of instruction. Google Search provides web pages that meet user needs in merely 12% of Chinese instances and 5% of English instances. We find **the search engine is better at processing short and keyword-like queries.**

Public information retrieval systems cannot access the private data with the university. These systems may fail to find correct evidence, potentially resulting in hallucinations during generation. Augmented by the hierarchical retrieval of campus-specific documents, ALOHA can yield accurate responses for 94% of Chinese instances, 87% for English and 78% for French, significantly surpassing ChatGPT and the search engines. Our system can also leverage the latest evidence to provide timely responses, while search engines usually return outdated web pages published before 2023.

---

[1] In this work, we employ the Programmable Search Engine API of Google to obtain results automatically.

| System | $\mathcal{K}$ | Accuracy | | | Timely |
|---|---|---|---|---|---|
| | | zh | en | fr | |
| ChatGPT | ✗ | 0.29 | 0.21 | 0.21 | ☹ |
| ERNIE Search | ✅ | 0.67 | 0.43 | 0.00 | ☹ |
| Google Search | ✅ | 0.12 | 0.05 | 0.01 | ☹ |
| ALOHA | ✅ | **0.94** | **0.87** | **0.78** | 😀 |

Table 3: Accuracy of the responses provided by each system in three distinct languages. $\mathcal{K}$ denotes whether the system can access external knowledge.

## 5 Case Study

We conduct case study to investigate ALOHA's performance in several key areas, including multilingual support, user-friendliness, capability to handle complex information, and timeliness.

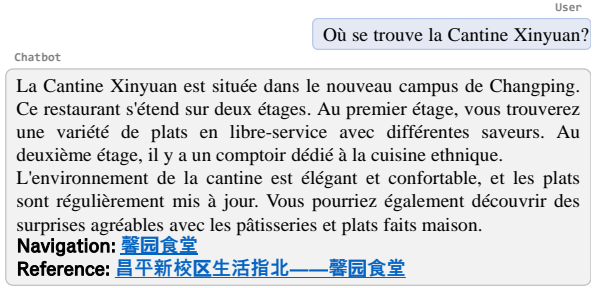

Figure 2: A case of location inquiry in French. See English translation in Appendix C.

### 5.1 Multilingual and User-Friendly Service

We hope that ALOHA can provide services to people using any language. It can accept non-Chinese inputs and generate responses in the same language. For example, a French-speaking user may inquire about the location of the Canteen Xinyuan by asking "*Où se trouve la Cantine Xinyuan?*" The response of our system is illustrated in Figure 2.

ALOHA yields fluent responses in French, providing users with the location of Canteen Xinyuan and detailed information about the services available on each floor. Our system also identifies *Canteen Xinyuan* is a building name, automatically providing a hyperlink to navigate to the location by external online map APPs. To prevent potential hallucinations of LLMs from misleading users, we further provide the original evidence as the **reference** for users to verify.

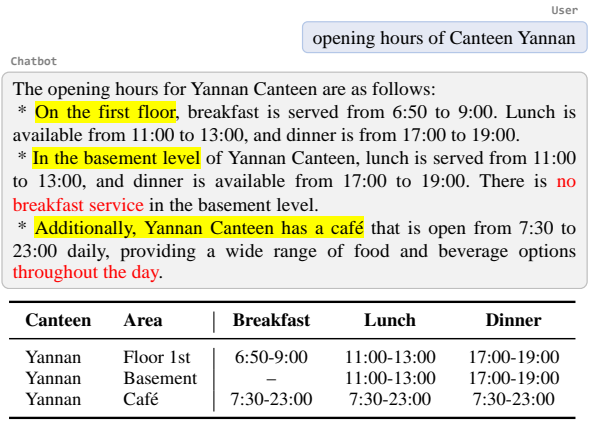

Figure 3: Inquiry case about opening hours, involving hierarchical retrieval and reasoning on complex tables.

### 5.2 Processing Complex Evidence

ALOHA frequently receives information consultations regarding schedules, reimbursement standards, and similar topics. Human users may feel confusing to identify critical information within complex tables and texts from a large number of documents. However, our system demonstrates strong capabilities to retrieve, aggregate, and analyze hierarchical semi-structured evidence.

Figure 3 shows an inquiry about opening hours and part of the tabular evidence. The user issues a vague query about the opening hours of Canteen Yannan. However, the canteen consists of three parts, whose opening hours are different. ALOHA can retrieve all rows about Canteen Yannan, providing a summarized response. We further find the system has promising capabilities to reason on tabular data. It can recognize the basement cannot provide breakfast and the café opens all the day, by comparing across table rows and aggregating cells with the same content.

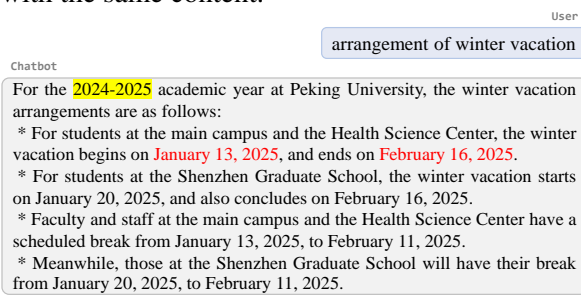

Figure 4: A case of inquiry about the next vacation.

### 5.3 Timeliness

Users often regard their current situation as common knowledge and omit this background information from the input. For example, when users inquiry about winter vacation schedules, they are typically asking about the next winter vacation. Figure 4 shows that ALOHA can adopt the latest information within retrieved documents, yielding a **timely** response of the arrangement of winter vacation in 2025. However, when asking Google Search (*winter vacation arrangement of Peking University*), it provides the outdated schedule of winter vacation in 2024. Although the search engine has crawled the latest web page, it still ranks the page in the third position.

### 5.4 Potential Hallucinations from Translation

To enhance the recall of retrieval, we translate all queries into Chinese. However, machine translation can occasionally introduce errors, which may subsequently propagate into the retrieval and generation processes. For example, when asking "*Where is the School of Electronic and Computer Engineering (ECE)?*", our system incorrectly responds that *the School of Electronics Engineer and Computer Engineer (EECS) is located at Beijing*. However, different from the school of EECS, ECE is actually a school located at the Shenzhen Campus. We note that the English name of ECE is not parallel to its Chinese name (, i.e., *School of Information Engineering*). Several words, especially the location names, do not have aligned literal meanings across different languages. Hence, the translated queries may occasionally fail to convey the user's original intent.

## 6 Deployment

Based on the ALOHA system, we develop the on-campus AI assistant *Xiaobei* for Peking University. This application has been integrated into the portal of Peking University and has provided consultation service around 33,000 times for 12,000 faculty, staff, and students, including more than 3,000 new undergraduate students.

## 7 Conclusion

In this work, we propose ALOHA, enhancing LLM agent with hierarchical retrieval to select evidence from extensive document with diverse structures and granularities. Our system performs better than existing chatbots or search engines to generate correct, timely and user-friendly responses in various languages. ALOHA facilitates campus-specific information retrieval and greatly benefits the orientation of new students. We hope ALOHA can inspire the development of more agents serving for education and public affairs.

# Ethics Statement

As a system constructed on LLM, ALOHA might yield a hallucinated output and mislead the users. We try our best to minimize potential harm of the AI system to users. We explicitly inform users that the agent can make mistakes. People should carefully examine and verify the important information during the conversations.

For the deployed application *Xiaobei*, we ensure that every user agrees to our recording of their access information before using this system.

We have implemented necessary safeguards through prompt engineering to prevent the system from providing harmful outputs. However, we must acknowledge that LLMs are not perfect in avoiding unfriendly generations. We kindly hope that this system can serve for campus-specific information retrieval rather than improper purposes.

Men, Ruize Gao, Runji Lin, Shijie Wang, Shuai Bai, Sinan Tan, Tianhang Zhu, Tianhao Li, Tianyu Liu, Wenbin Ge, Xiaodong Deng, Xiaohuan Zhou, Xingzhang Ren, Xinyu Zhang, Xipin Wei, Xuancheng Ren, Xuejing Liu, Yang Fan, Yang Yao, Yichang Zhang, Yu Wan, Yunfei Chu, Yuqiong Liu, Zeyu Cui, Zhenru Zhang, Zhifang Guo, and Zhihao Fan. 2024. Qwen2 technical report.

## A  Interaction with External APIs

Following previous work (Hsieh et al., 2023), we regard the external APIs as tools and prompt the LLM to reason as a planner to determine the invokation of these APIs. We provide LLM with the functions and the formats of received parameters (if needed) of each tool. For example, the function of an online map application is to provide precise location information and navigation service. The online map application should receive a location name as input.

We prompt the LLM to determine whether to use a tool according to the response. If the response contains a location name, the LLM can recognize it and select the online map applications. Our system then tries to invoke these applications with the location name as input. If the invokation is valid to be executed, we will integrate these applications into the interface and demonstrate them to the user. Figure 5 demonstrates an example that the user asks about the location of Canteen Xinyuan. The response contains a hyperlink to online map applications with the location name as input. Users can conveniently turn to the navigation service.

We also engage several campus-sepcific APIs, such as *Inquiry of Available Classrooms* and *Busy Index of Canteen*. Figure 6 illustrates the query about the least crowded canteen. Our system identifies the user's intent and provides the hyperlink to the page of *Busy Index of Canteen*.

## B  Details of Intent Classification

### B.1  Intent Classes

In this work, we find the user queries about working schedule, reimbursement, and standards of expense always need to retrieve evidence from semi-structured documents. Therefore, we empirically propose to train a classifier to determine whether a query contains an intent mentioned above. We further categorize these intents into 11 classes, which are detailed as following:

- Routine Reimbursement
- Reimbursement for Software Development or Purchase
- Inter-City Transportation Expense
- International Transportation Expense
- Accommodation Expense
- Field Investigation Expense
- Conference Expense
- Expert Consultation Expense
- Service Expense for Off-Campus Personnel
- Opening Schedule of Buildings
- Service Schedule of Buildings during Holiday Period

### B.2  Data Augmentation

Through user study, we find the queries in real world may have different patterns from our seed questions. Users usually add their identity information; for example, "*I am a full professor. Can I reimburse first-class train tickets?*" They also often provide colloquial queries like "*I wanna have my money cost for the trip of Kunming back.*" Such queries can become out-of-distribution instances compared to the seed questions.

To address these challenges, we employ `gpt-4o` to paraphrase the seed questions. We prompt the LLM to modify the phrasing of the questions, without changing their original intents. We allow LLM to paraphrase a question based on following rules:

1. To simplify the given question.
2. To add the identity information of imaginary faculty, staff, and students.
3. To change the order of words without altering the original meaning.
4. To rewrite as a synonymous question.

## C  English Translation of Case in French

We here provide the English translation of the case in Figure 2.

**User:** Where is Canteen Xinyuan?

**Chatbot:** Canteen Xinyuan is located in the new Changping Campus. This restaurant is spread over two floors. On the first floor you will find a variety of self-service dishes with different flavors. On the second floor, there is a counter dedicated to ethnic cuisine.

The canteen environment is elegant and comfortable, and the dishes are regularly updated. You

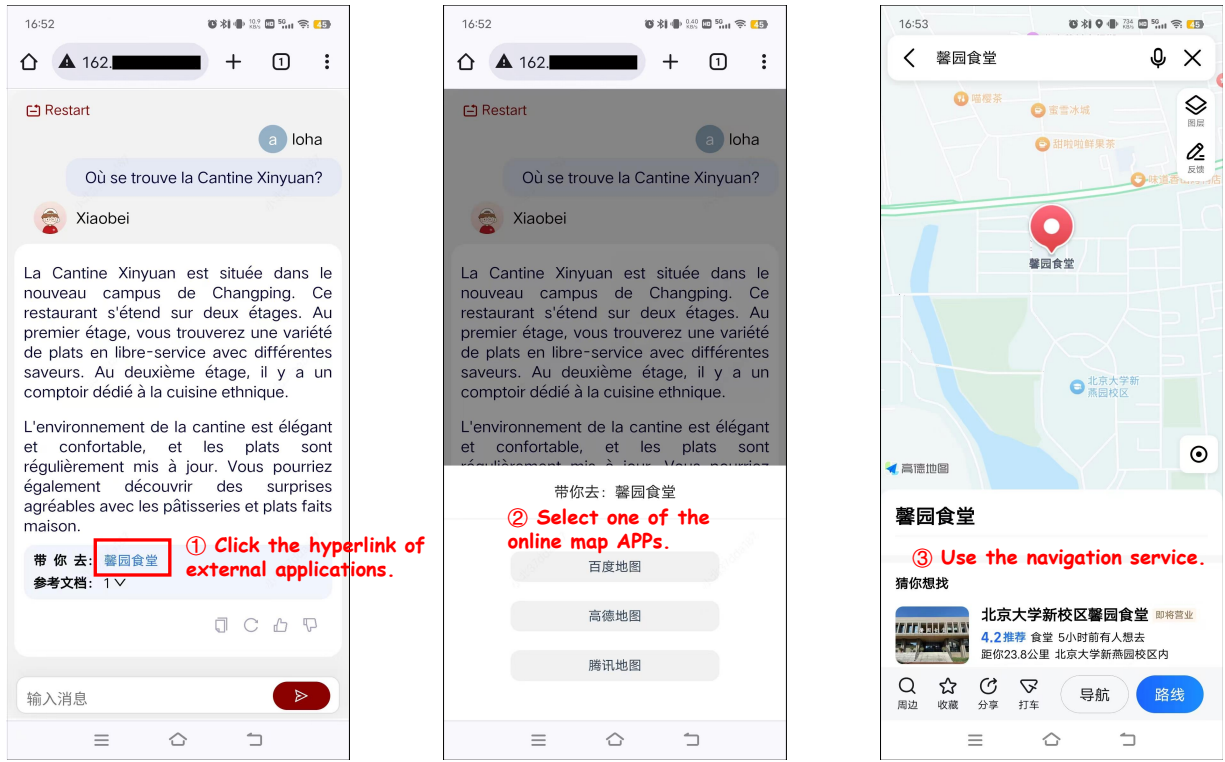

Figure 5: Demonstration of our system when integrating the online map applications into the interface.

might also discover pleasant surprises with the homemade pastries and dishes.

    **Navigation:** Canteen Xinyuan

    **Reference:** Guide of the life in the new Changping Campus – Canteen Xinyuan

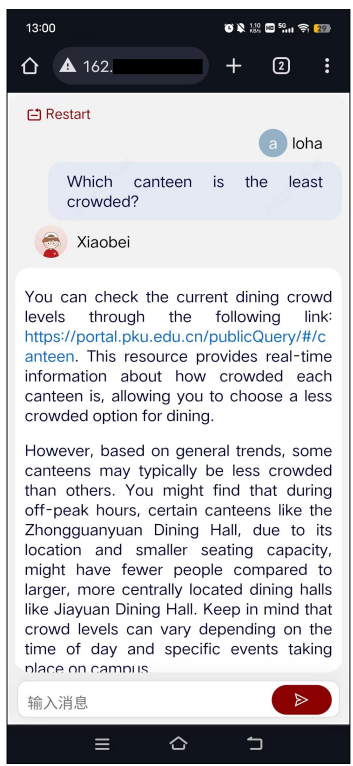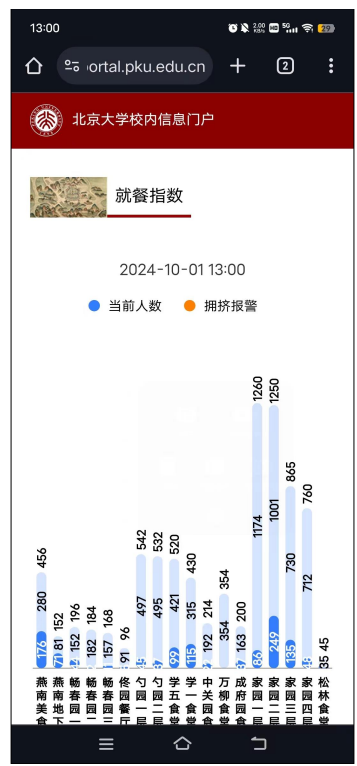

Figure 6: Demonstration of the query about not-crowded canteen.